\documentclass[twocolumn, switch]{article} 

\usepackage{preprint}

\usepackage{enumitem}
\usepackage{amsmath, amsthm, amssymb, amsfonts}
\usepackage{bm}
\usepackage{amsmath}
\usepackage{graphicx}
\usepackage{subfigure}
\usepackage{adjustbox}
\usepackage[algoruled,algo2e]{algorithm2e}

\usepackage{setspace}
\usepackage{amssymb}
\usepackage{booktabs}
\usepackage{array}
\usepackage{color}
\usepackage{multirow}
\usepackage{multicol}
\usepackage{threeparttable}
\usepackage{url}
\usepackage[page]{appendix}
\usepackage{makecell}

\usepackage[numbers,square]{natbib}
\usepackage[utf8]{inputenc}	
\usepackage[T1]{fontenc}	
\usepackage{xcolor}		
\usepackage[colorlinks = true,
            linkcolor = blue,
            urlcolor  = blue,
            citecolor = blue,
            anchorcolor = black]{hyperref}	
\usepackage{booktabs} 		
\usepackage{nicefrac}		
\usepackage{microtype}		
\usepackage{lineno}		
\usepackage{float}			

\usepackage{newfloat}
\usepackage{arydshln}
\DeclareFloatingEnvironment[name={Supplementary Figure}]{suppfigure}
\usepackage{sidecap}
\sidecaptionvpos{figure}{c}

\usepackage{titlesec}
\titlespacing\section{0pt}{12pt plus 3pt minus 3pt}{1pt plus 1pt minus 1pt}
\titlespacing\subsection{0pt}{10pt plus 3pt minus 3pt}{1pt plus 1pt minus 1pt}
\titlespacing\subsubsection{0pt}{8pt plus 3pt minus 3pt}{1pt plus 1pt minus 1pt}

\usepackage{graphics}
\usepackage{hyperref}
\usepackage{color}
\usepackage{multirow}
\usepackage{hhline}
\usepackage{subfigure}
\usepackage{lipsum}
\usepackage{flushend}
\usepackage{dblfloatfix}

\usepackage[table]{xcolor}
\definecolor{mygray}{gray}{.9}
\usepackage{colortbl}

\title{SemPT: Semantic Prompt Tuning for Vision-Language Models}

\usepackage{authblk}

\author[1]{Xiao Shi}
\author[1]{Yangjun Ou*}
\author[2]{Zhenzhong Chen}
\affil[1]{School of Computer Science and Artificial Intelligence, Wuhan Textile University}
\affil[2]{School of Remote Sensing and Information Engineering, Wuhan University}

\begin{document}

\twocolumn[ 
  \begin{@twocolumnfalse} 
  
\maketitle

\begin{abstract}

Visual transfer learning for unseen categories presents an active research topic yet a challenging task, due to the inherent conflict between preserving category-specific representations and acquiring transferable knowledge. Vision-Language Models (VLMs) pre-trained on large amounts of image-text pairs offer a promising solution. However, existing prompt tuning methods rely on sparse category labels or disparate LLM-generated descriptions, which fragment knowledge representation and hinder transferability. To address this limitation, we introduce Semantic Prompt Tuning (SemPT), a novel framework that tackles the generalization challenge by leveraging shared attribute-level knowledge across categories. Specifically, SemPT adopts a two-step prompting strategy to guide LLM in extracting shared visual attributes and generating attribute-level descriptions, capturing transferable semantic cues beyond labels while ensuring coherent structure. Then, visually guided weighting is applied to the embeddings of attribute-level descriptions to reduce noise from irrelevant attributes and enhance the text embeddings. Additionally, image embeddings are jointly aligned with both label and attribute-enhanced text embeddings, balancing discrimination for seen categories and transferability to unseen ones. Considering the availability of category exposure, our inference dynamically selects between standard label embeddings for seen categories and attribute-enhanced embeddings for unseen ones to ensure effective adaptation. Extensive experiments on 15 benchmark datasets demonstrate that SemPT achieves state-of-the-art performance across various settings, including base-to-novel generalization, cross-dataset transfer, cross-domain transfer, and few-shot learning.

\end{abstract}

\vspace{0.4cm}

  \end{@twocolumnfalse} 
] 

\newcommand\blfootnote[1]{%
\begingroup
\renewcommand\thefootnote{}\footnote{#1}%
\addtocounter{footnote}{-1}%
\endgroup
}

\section{Introduction}

{\blfootnote{*Corresponding author: Yangjun Ou, E-mail:yjou@wtu.edu.cn}}

\begin{figure*}[t]
  \centering
  \includegraphics[width=0.95\textwidth]{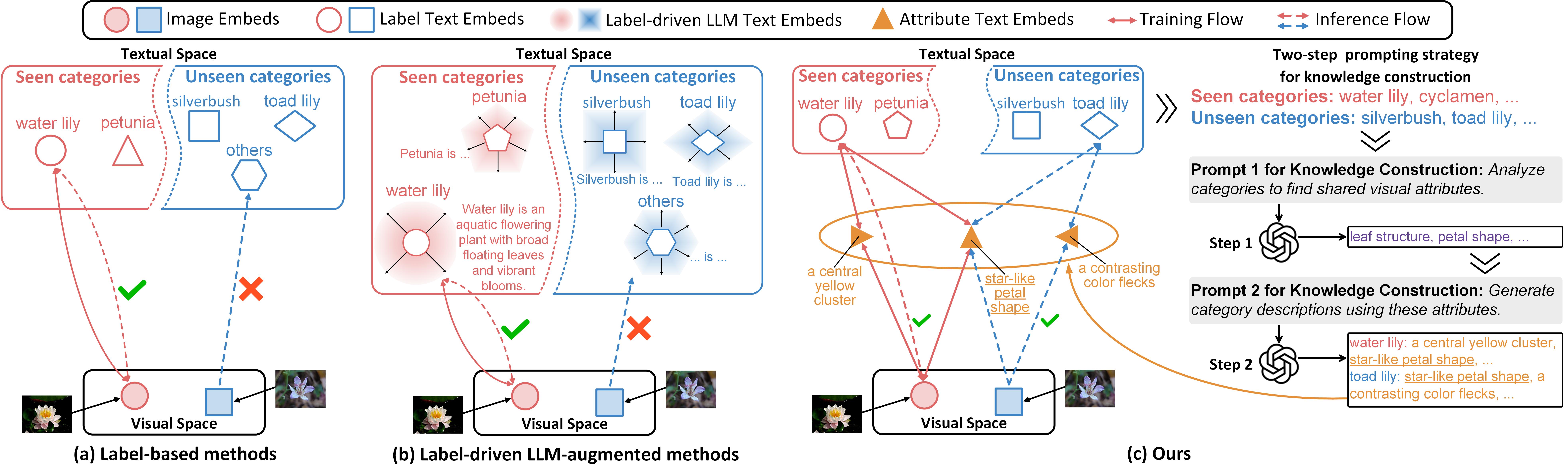}
  \caption{Illustration of existing methods and the proposed SemPT framework. (a) Label-based methods rely on sparse category labels without attribute modeling, preventing knowledge transfer from seen to unseen categories. (b) Label-driven LLM-augmented methods generate descriptions to expand semantics, but insufficient attribute decomposition leads to fragmented knowledge representation. (c) The proposed SemPT framework explicitly models shared attribute-level knowledge through a two-step prompting strategy, significantly improving the generalization of VLMs.}
\label{fig:motivation}
\end{figure*}
 
Visual-Language Models (VLMs), such as CLIP~\cite{CLIP}, are pre-trained through contrastive learning on large-scale image-text pairs, forming a foundational knowledge base for transfer learning tasks. However, a fundamental conflict exists in transfer learning between preserving category-specific representations for seen categories and acquiring transferable knowledge for unseen categories. To enhance the transfer capabilities of VLMs, two primary approaches have emerged, i.e., full fine-tuning~\cite{ViFi-CLIP,FD-CLIP,CityLLaVA} and prompt tuning~\cite{CoOp,CoCoOp,MaPLe,PromptSRC}. Among these, prompt tuning methods can achieve performance parity with, or even outperform, fully fine-tuning methods with few learnable parameters.

Existing prompt tuning methods primarily adopt two kinds of strategies for textual representation. Label-based methods~\cite{MetaPrompt,MMA,KIM,PromptKD} learn prompts solely from category names, efficiently optimizing embeddings, but lack explicit attribute semantics for transfer learning. While label-driven LLM-augmented methods ~\cite{HPT,CoPrompt,ArGue,CoCoLe} employ LLM to generate richer descriptions and enhance semantic coverage, they neglect to model shared attributes across categories. Due to insufficient attribute decomposition, the generated descriptions are diverse yet dispersed, leading to fragmented text embeddings.

Such fragmentation inherently undermines the transferability of learned knowledge. The model learns to align image embeddings with textual embeddings from sparse labels (e.g., ``\textit{water lily}'' and ``\textit{petunia}'' in Fig.~\ref{fig:motivation}(a)) or disparate descriptions (e.g., ``\textit{Water lily is an aquatic flowering plant with broad floating leaves and vibrant blooms.}'' in Fig.~\ref{fig:motivation}(b)) during training. However, the fragmented text embeddings fail to establish meaningful semantic connections across categories (e.g., from ``\textit{water lily}'' to ``\textit{toad lily}''), preventing the model from generalising to unseen categories during inference. 
This gap reveals that effective transfer learning needs to encode cross-category shared attributes as semantic bridges, 
Enabling the model to decompose unseen categories into combinations of previously learned attributes.

In this paper, we construct transferable semantic pathways from seen to unseen categories by leveraging shared attribute-level knowledge. As illustrated in Fig.~\ref{fig:motivation}(c), this approach transforms a fragmented text-embedding landscape into a semantically connected space grounded in shared visual attributes. Shared attributes (e.g., ``\textit{leaf structure}'', ``\textit{petal shape}'') are extracted and used to generate attribute-level descriptions for each category (e.g., ``\textit{water lily: floating leaf pads, star-like petal shape}'', ``\textit{toad lily: star-like petal shape, glossy leaf finish}''). The shared attribute “\textit{star-like petal shape}” forms an explicit semantic link between categories (e.g., from ``\textit{water lily}'' to ``\textit{toad lily}''), enabling knowledge transfer across the seen–unseen boundary, where labels anchor seen categories and attributes guide unseen ones.

To implement this approach, we propose Semantic Prompt Tuning (SemPT), which consists of four key components. First, a two-step prompting strategy guides the LLM to construct an attribute-level knowledge space: LLM initially extracts shared visual attributes across all categories, then generates attribute-level descriptions grounded in these attributes. Second, the attribute-level descriptions are weighted according to visual relevance, then fused with category label text embeddings to form attribute-enhanced text embeddings. Third, a dual-branch supervision scheme aligns image embeddings with both label and attribute-enhanced text embeddings, balancing discrimination and transferability. Finally, inference is performed based on category exposure: label text embeddings are applied to seen categories, with attribute-enhanced text embeddings handling unseen ones, completing training-to-inference bridging for robust recognition.

To validate our approach, we conduct extensive experiments across challenging evaluation settings, including base-to-novel generalization, cross-dataset transfer, cross-domain transfer, and few-shot learning. These diverse scenarios thoroughly test generalization across different categories, datasets, domains, and data conditions. SemPT consistently achieves state-of-the-art performance across all settings and exhibits broad compatibility with various VLM architectures, establishing its versatility as a general enhancement strategy.

The main contributions of this paper are as follows:

\begin{itemize}
\item Semantic Prompt Tuning (SemPT) is proposed to tackle the generalization challenge in transfer learning by leveraging shared attribute-level knowledge, resolving fragmented knowledge representation, and enabling robust cross-category understanding.

\item A two-step prompting strategy constructs attribute-level textual descriptions, which establish transferable semantic pathways for progressively augmenting text embeddings through dedicated modules, significantly strengthening knowledge transfer capabilities.

\item A unified training-inference adaptation mechanism is proposed to tailor text representations to category exposure, effectively balancing generalization and specificity.

\item We conduct extensive experiments on 15 benchmark datasets, demonstrating that SemPT achieves state-of-the-art results across all evaluation settings while maintaining excellent compatibility with various baselines.
\end{itemize}

The rest of this paper is organized as follows: Section II reviews related work that lays the foundation for our study. Section III presents the proposed SemPT framework in detail, including its four core modules. Section IV describes the experimental setup and provides extensive experimental results with detailed analysis. Section V concludes the paper.

\section{Related Work}

\subsection{Vision-Language Models}

Vision-Language Models (VLMs) enable joint understanding of visual and textual modalities by aligning their representations in a shared embedding space. Most VLMs adopt contrastive learning as the core training objective, encouraging similarity between matched image-text pairs while discouraging mismatched ones. 

CLIP~\cite{CLIP} is a seminal work that aligns visual and textual representations using a dual-encoder framework trained on 400 million image-text pairs, achieving strong zero-shot performance on various vision benchmarks. ALIGN~\cite{ALIGN} scales this paradigm by using noisy alt-text annotations and a larger dataset, demonstrating that simple architectures can benefit significantly from massive pre-training. BLIP~\cite{BLIP} introduces a new direction by combining vision-language understanding and generation in a unified pre-training pipeline, using a bootstrapped learning mechanism to improve robustness under weak supervision.

Recent efforts have focused on enhancing architectural capacity and improving interactions between modalities. For example, BLIP-2~\cite{BLIP-2} introduces a Q-Former with learnable queries to extract relevant visual features better, while Flamingo~\cite{Flamingo} employs a Perceiver Resampler to facilitate cross-modal selection. ViLA~\cite{ViLA} investigates more efficient pre-training strategies to boost training efficiency and downstream transfer performance. Building on these advances, MSGM~\cite{MSGM} presents a vision-language pre-training method designed to enhance zero-shot generalization in embodied reasoning tasks.

These advances have greatly improved VLMs' generalization, enabling applications such as cross-modal retrieval, few-shot learning, and compositional reasoning. However, most existing VLMs still rely on global supervision, which limits their adaptability to new domains, especially in low-resource or domain-shifted settings.

\subsection{Label-based Prompt Tuning}

Prompt tuning was first proposed in natural language processing (NLP) as a parameter-efficient alternative to full fine-tuning, enabling pre-trained language models to adapt to new tasks via learnable and task-specific prompts~\cite{PromptTuning}. This idea was later extended to VLMs, where CoOp~\cite{CoOp} pioneered the use of continuous prompts to replace hand-crafted templates for zero-shot image classification. While effective, CoOp relies on static prompts that lack input awareness, limiting its generalization to unseen domains.

To address CoOp’s limited generalization, subsequent methods have focused on dynamic and multi-level prompt adaptation as well as architectural innovations to enhance image-text alignment. 
CoCoOp~\cite{CoCoOp} introduces dynamic prompts conditioned on each input image, improving transferability but limited to image-based conditioning without textual adaptation. MaPLe~\cite{MaPLe} extends the idea by jointly learning prompts for both the visual and textual branches. It applies multi-layer prompt tuning across transformer blocks. Cross-modal coupling is enforced throughout the network, enabling deeper interactions beyond the final output layer. Beyond prompt design, MMA~\cite{MMA} and MMRL~\cite{MMRL} propose architectural improvements: MMA uses lightweight adapters to aggregate features into a shared space, enabling effective gradient flow, while MMRL builds a modality-agnostic token space for fine-grained token-level interaction. In a similar vein, TIPPLE~\cite{TIPPLE} proposes test-time task-to-instance prompt learning to adapt vision-language models dynamically. At the same time, KIM~\cite{KIM} explores knowledge injection strategies to enhance model adaptability and robustness.

Another line of work focuses on enhancing the generalization of prompt tuning under distribution shifts or low-resource conditions. 
PromptSRC~\cite{PromptSRC} regularizes prompt learning through mutual agreement constraints and prompt ensembling to improve generalization and stability, while employing diversity regularization to reduce overfitting. ADAPT~\cite{ADAPT} integrates adversarial training with prompt tuning, proposing adversarial dual prompts to address unsupervised domain adaptation and improve robustness against domain shifts. MetaPrompt~\cite{MetaPrompt} addresses domain shift by learning domain-invariant prompts via meta-learning, training prompts to generalize across held-out domains through episodic training. PromptKD~\cite{PromptKD} adopts a transductive zero-shot learning framework that distills knowledge from a teacher model into prompts using unlabeled target-domain data, aligning student predictions with teacher outputs for unseen categories without supervision. 

These methods provide an efficient foundation for adapting VLMs, but often lack explicit modeling of shared semantic structures, which limits their interpretability and generalization ability, especially under distribution shifts and scarce supervision.

\subsection{LLM-augmented Prompt Tuning}

The rapid advancement of Large Language Models (LLMs), such as ChatGPT~\cite{ChatGPT}, LLaMA~\cite{LLaMA}, Gemini~\cite{Gemini}, and Claude~\cite{Claude}, has revolutionized natural language understanding and generation. These capabilities have inspired the integration of LLMs into VLMs to improve prompt tuning, leveraging their linguistic knowledge and reasoning abilities for more informative and transferable prompts.

Recent LLM-augmented prompt tuning methods improve prompt quality and model adaptability through various strategies. TCA~\cite{TCA} learns prompt-enhanced context features to better model weakly-supervised video anomaly detection scenarios. To capture richer semantic structures, HPT~\cite{HPT} prompts LLMs to generate hierarchical category descriptions that encode multi-level relationships, enhancing generalization to unseen categories. 
CoPrompt~\cite{CoPrompt} mitigates overfitting by enforcing consistency between trainable prompts and LLM-generated features, aligning learned prompts with linguistic priors. 
CoCoLe~\cite{CoCoLe} distills high-level conceptual knowledge from LLMs into a discrete codebook space, providing semantically grounded tokens that facilitate robust cross-domain generalization. ArGue~\cite{ArGue} shifts focus to fine-grained attribute-level semantics, injecting discriminative descriptions into prompts to enhance subtle visual differentiation. ProText~\cite{ProText} takes a further step by supervising prompt learning purely through LLM-generated text, enabling zero-shot transfer without requiring labeled images.

These LLM-augmented methods advance prompt learning by enriching semantic representations with external linguistic knowledge, enhancing adaptability and generalization in VLMs. Nevertheless, their limited focus on modeling shared attribute-level semantics results in fragmented text embeddings that hinder knowledge transfer and degrade robustness under distribution shifts.

\begin{figure*}[t]
  \centering
  \includegraphics[width=0.95\textwidth]{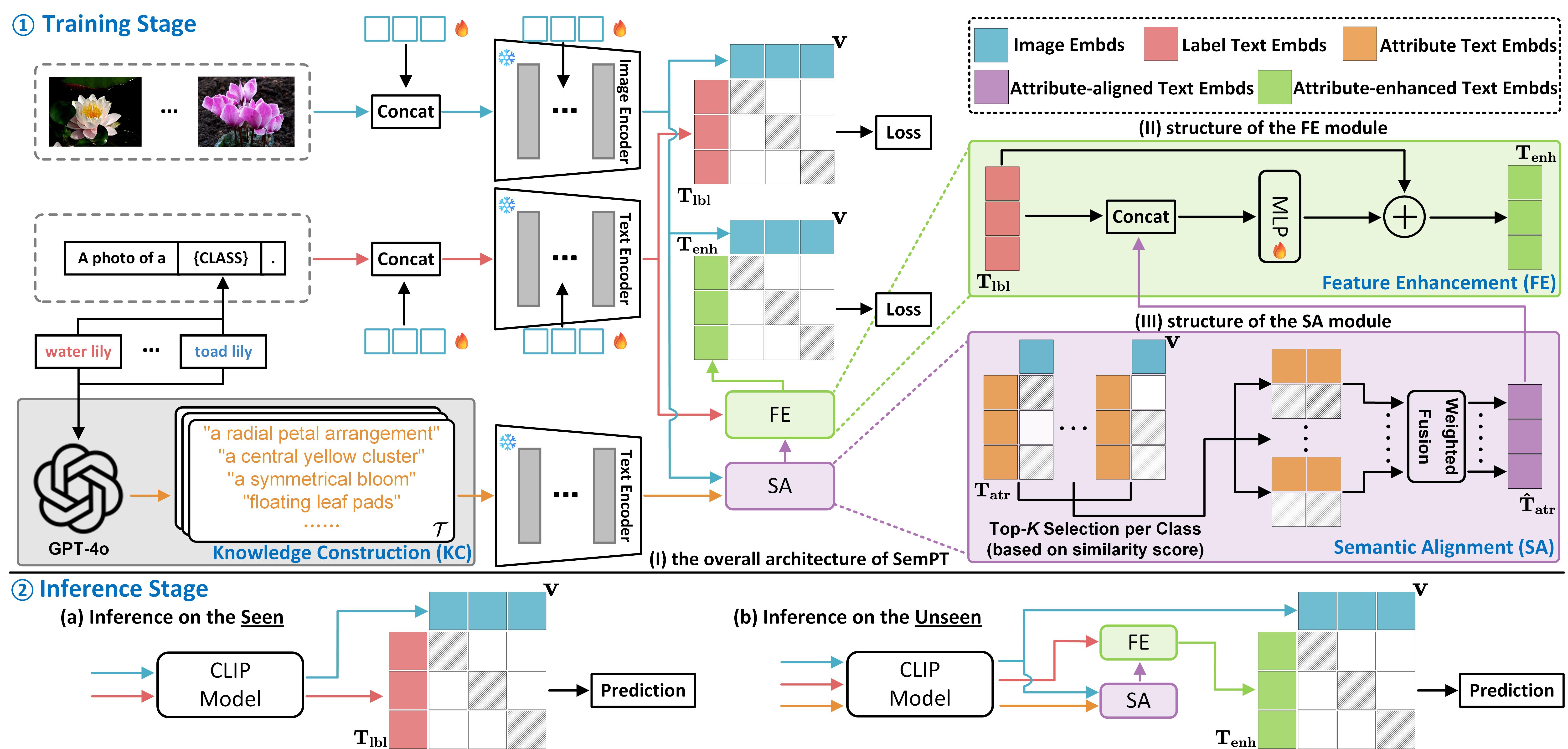}
  \caption{Overview of the SemPT framework. 
\textbf{Training Stage}: 
In the KC module, GPT-4o is used in a two-step prompting strategy to extract shared visual attributes and generate attribute-level descriptions $\mathcal{T}$ across all categories.
The SA module selects the top-$K$ most visually relevant descriptions per category based on image-text similarity, followed by softmax-weighted averaging to obtain attribute-aligned text embeddings $\hat{\text{T}}_{\text{atr}}^{(i)}$. 
The FE module integrates label text embeddings $\text{T}_{\text{lbl}}$ and attribute-aligned text embeddings $\hat{\text{T}}_{\text{atr}}$ via concatenation, a linear projection, and a residual connection, yielding attribute-enhanced text embeddings $\text{T}_{\text{enh}}$. 
Both $\text{T}_{\text{lbl}}$ and $\text{T}_{\text{enh}}$  are supervised to facilitate robust recognition.
\textbf{Inference Stage}: Label text embeddings $\text{T}_{\text{lbl}}$ are used for \underline{seen} categories, and attribute-enhanced text embeddings $\text{T}_{\text{enh}}$ for \underline{unseen} ones.}
  \label{fig:framework}
\end{figure*}

\section{Methodology}

To tackle the generalization challenge caused by fragmented knowledge representation, we propose Semantic Prompt Tuning (SemPT) for diverse transfer scenarios. The overall architecture of the proposed SemPT is illustrated in Fig.~\ref{fig:framework}. 

\subsection{Overview}

Given a dataset $\mathcal{D} = \{(x_j, y_j)\}_{j=1}^M$ consisting of images $x_j$ and their corresponding labels $y_j$, where labels are drawn from either seen categories $\mathcal{C}_{s} = \{c_i\}_{i=1}^{N_s}$ or unseen categories $\mathcal{C}_{u} = \{c_i\}_{i=1}^{N_u}$, our goal is to adapt a CLIP-based vision-language model that enables accurate recognition across both $\mathcal{C}_s$ and $\mathcal{C}_u$ during inference.

To this end, the proposed SemPT framework comprises four key modules: Knowledge Construction (KC), Semantic Alignment (SA), Feature Enhancement (FE), and Unified Training-Inference Adaptation (UTIA).

\textbf{Knowledge Construction (KC).}  The KC module extracts shared visual attributes and generates attribute-level textual descriptions for all categories via a two-step prompting strategy using LLM.  
In the first step, the LLM identifies a set of shared visual attributes across all categories, forming a unified semantic vocabulary $\mathcal{A} = \{a_m\}_{m=1}^{M_a}$.  
In the second step, based on $\mathcal{A}$, LLM generates a set of attribute-level descriptions:  $\mathcal{T} = \{ t_i^j \mid \substack{ j=1,\ldots,S \\ i=1,\ldots,N } \}$, where $N = N_s + N_u$ is the total number of categories, $S$ is the number of descriptions per category, and $t_i^j$ denotes the $j$-th attribute-level description for category $c_i \in \mathcal{C}_{s} \cup \mathcal{C}_{u}$.

\textbf{Semantic Alignment (SA).}  
The SA module selects and aggregates the most visually relevant attribute-level descriptions for each input image.  
Given an image $x_j$, we obtain its embedding $\text{v} \in \mathbb{R}^d$ via the image encoder $E_V$, where $d$ is the embedding dimension.  
We compute cosine similarities between $\text{v}$ and all attribute text embeddings $\text{T}_{\text{atr}} \in \mathbb{R}^{N \times S \times d}$, derived by encoding attribute-level descriptions $\mathcal{T}$ via the text encoder.  
For each category, the top-$K$ most relevant descriptions are selected and aggregated with softmax-based weighting, resulting in attribute-aligned embeddings $\hat{\text{T}}_{\text{atr}} \in \mathbb{R}^{ N\times d}$.

\textbf{Feature Enhancement (FE).} The FE module fuses label text embeddings with attribute-level semantics.
The label text embeddings $\text{T}_{\text{lbl}} \in \mathbb{R}^{N \ N\times d}$ are obtained by encoding each category name from $\mathcal{C}_s$ and $\mathcal{C}_u$ via the text encoder. These embeddings are then combined with $\hat{\text{T}}_{\text{atr}}$ through a projection layer, followed by a residual connection. This results in attribute-enhanced text embeddings $\text{T}_{\text{enh}} \in \mathbb{R}^{N \times d}$.

\textbf{Unified Training-Inference Adaptation (UTIA).}  
The UTIA module employs a dual-embedding strategy to bridge the gap between training and inference.  
During training, $\text{T}_{\text{lbl}}$ and $\text{T}_{\text{enh}}$ are supervised by separate classification losses, which are jointly optimized via a weighted sum.  
During inference, $\text{T}_{\text{lbl}}$ is used for seen categories to preserve discriminative power, while $\text{T}_{\text{enh}}$ is applied to unseen categories to enhance generalization through attribute-level semantics.

\begin{figure*}[t]
  \centering
  \includegraphics[width=0.98\linewidth]{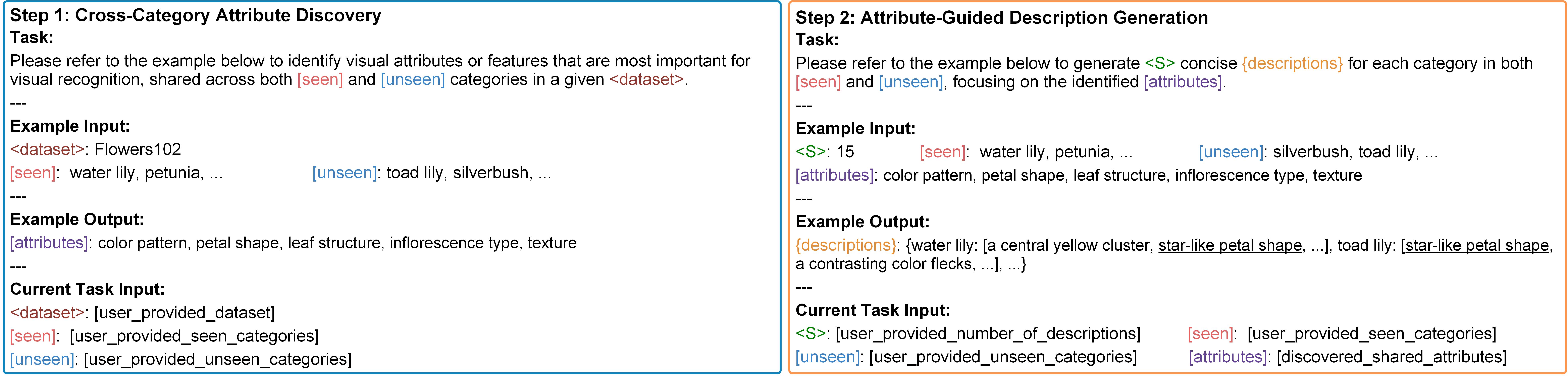}
  \caption{Two-step prompting strategy for knowledge construction using LLM. Step 1 identifies diverse and complementary visual attributes shared across seen and unseen categories. Step 2 generates low-redundancy, attribute-guided descriptions for each category, providing a comprehensive semantic basis for subsequent selection of informative textual features aligned with visual representations.}
  \label{fig:llm}
\end{figure*}

\subsection{Knowledge Construction}

The foundation of the SemPT framework lies in constructing transferable semantic representations that bridge seen and unseen categories. To this end, we propose a two-step prompting strategy using LLM to generate structured and low-redundancy textual descriptions grounded in shared visual attributes. These descriptions provide a unified and discriminative semantic basis that facilitates the alignment of visual and textual features and supports effective knowledge transfer across categories, as illustrated in Fig.~\ref{fig:llm}.

\textbf{Step 1: Cross-Category Attribute Discovery.}  
To construct a shared semantic foundation, we first prompt the LLM to identify a set of visual attributes broadly applicable to both seen ($\mathcal{C}_s$) and unseen ($\mathcal{C}_u$) categories.  
Specifically, we design a prompt template that instructs the LLM to extract visual attributes covering diverse semantic dimensions and functional characteristics.
The prompt emphasizes diversity and orthogonality across visual dimensions to ensure comprehensive semantic coverage and minimize attribute redundancy.  
To enhance the quality of extracted attributes, we incorporate category examples from both $\mathcal{C}_s$ and $\mathcal{C}_u$ in the prompt, enabling the LLM to identify attributes that are generalizable across the category spectrum.
This process produces a shared attribute vocabulary $\mathcal{A}$, which serves as the semantic backbone guiding subsequent description generation.

\textbf{Step 2: Attribute-Guided Description Generation.}  
Based on $\mathcal{A}$, we then prompt the LLM to generate multiple concise descriptions for each category $c_i \in \mathcal{C}_s \cup \mathcal{C}_u$.  
For each category $c_i$, we prompt the LLM to create $S$ distinct descriptions, where each description highlights a different combination of relevant attributes from $\mathcal{A}$.
Each description references a subset of shared attributes, ensuring semantic consistency while capturing category-specific visual traits.  
To maintain description diversity, we employ varied prompt formulations and explicitly instruct the LLM to avoid repetitive phrasing across descriptions for the same category.
Furthermore, we constrain each description to be concise to facilitate efficient processing while preserving semantic richness.
This results in a structured set of attribute-level descriptions $\mathcal{T}$.

By grounding descriptions in the shared vocabulary $\mathcal{A}$, we ensure semantic alignment across categories while preserving distinctiveness. These low-redundancy and attribute-aware descriptions $\mathcal{T}$ serve as transferable semantic anchors to facilitate alignment with visual representations in downstream modules, enabling effective knowledge transfer from seen to unseen categories.

\subsection{Semantic Alignment}

While attribute-level descriptions $\mathcal{T}$ offer semantically rich textual knowledge, not all are equally aligned with visual content. Among the multiple descriptions per category, some exhibit strong visual grounding, while others may be redundant, overly abstract, or weakly correlated with visual representations. To mitigate semantic noise and enhance the relevance of textual supervision, we propose a dynamic semantic alignment mechanism that adaptively emphasizes the most visually aligned descriptions for each category.

Each description $t_i^j \in \mathcal{T}$ is encoded using the text encoder $E_T$, resulting in normalized attribute text embeddings $\text{T}_{\text{atr}}$:
\begin{equation}
\text{T}_{\text{atr}} = \{ E_T(t_i^j) \}_{i=1,j=1}^{(N,S)} \in \mathbb{R}^{N \times S \times d}
\end{equation}

This set of attribute text embeddings serves as a semantic reservoir at the category level, capturing diverse linguistic attributes that provide complementary signals beyond the original category names.

Given an input image $x_j$, the image encoder $E_V$, augmented with a learnable visual prompt $P_V$, produces a normalized image embedding $\text{v}$:
\begin{equation}
\mathbf{v} = E_V(x_j \oplus P_V) \in \mathbb{R}^d
\end{equation}
where $\oplus$ denotes the prompt integration operator that prepends the learnable prompt tokens to the input.

To assess the visual relevance of each textual description $t_i^j$ associated with category $c_i$, we compute the cosine similarity between the image embedding $\text{v}$ and the corresponding attribute text embedding $\text{T}_{\text{atr}}^{(i,j)}$:
\begin{equation}
s_{i,j} = \langle \text{v}, \text{T}_{\text{atr}}^{(i,j)} \rangle, \quad j = 1, 2, \ldots, S
\end{equation}
where $\langle \cdot, \cdot \rangle$ denotes the cosine similarity function.

Based on the similarity scores, we identify the top-$K$ most relevant descriptions for each category $c_i$.  
Let $\Omega_{c_i} = \{j_1, j_2, \ldots, j_K\}$ denote the index set corresponding to the $K$ highest similarity scores for category $c_i$, where $K < S$.  
To further reflect the relative importance of each selected description, we compute attention weights over $\Omega_{c_i}$ using a temperature-scaled softmax function:
\begin{equation}
w_{i,j} = \frac{\exp(s_{i,j}/\tau)}{\sum_{j' \in \Omega_{c_i}} \exp(s_{i,j'}/\tau)}, \quad j \in \Omega_{c_i}
\end{equation}
where $\tau$ is a temperature parameter controlling the sharpness of the attention distribution.

The final semantically aligned textual representation for category $c_i$ is then obtained as a weighted combination of the selected attribute text embeddings:
\begin{equation}
{\hat{\text{T}}^{(i)}_{\text{atr}}} = \sum_{j \in \Omega_{c_i}} w_{i,j} \cdot \text{T}_{\text{atr}}^{(i,j)}
\end{equation}

This process is repeated for all categories to construct the attribute-aligned text embeddings $\hat{\text{T}}_{\text{atr}}$:
\begin{equation}
{\hat{\text{T}}}_{\text{atr}} = [\hat{\text{T}}_{\text{atr}}^{(1)}, \hat{\text{T}}_{\text{atr}}^{(2)}, \ldots, \hat{\text{T}}_{\text{atr}}^{(N)}] \in \mathbb{R}^{N \times d}
\end{equation}

This set of attribute-aligned embeddings serves as the final semantic representation for each category, integrating visually grounded and semantically informative signals distilled from diverse candidate descriptions.

By incorporating both top-$K$ selection and attention-based weighting, the proposed alignment mechanism ensures that the final attribute-aligned text embeddings $\hat{\text{T}}_{\text{atr}}$ are semantically informative and visually grounded. This facilitates more robust visual-semantic alignment, particularly in open-set scenarios where shared attributes play a crucial role in recognizing unseen categories. Moreover, the use of adaptive alignment mitigates over-reliance on any single description, promoting generalization and reducing semantic bias arising from the diversity and inconsistency of attribute-level descriptions $\mathcal{T}$.

\subsection{Feature Enhancement}

To further enhance category representations, we build on the semantically attribute-aligned text embeddings by introducing a feature enhancement mechanism that constructs more expressive and generalizable category representations. While the attribute-level descriptions $\mathcal{T}$ encapsulate fine-grained, visually grounded semantics that are beneficial for cross-category generalization, the label text embeddings derived from category names offer concise and semantically coherent anchors that reflect the canonical identity of each class. To exploit the complementary strengths of these two representations, we propose an adaptive integration strategy that enriches the label text embeddings by integrating them with fine-grained semantic attributes derived from attribute-level descriptions $\mathcal{T}$.

Let $\mathcal{C} = \{c_i\}_{i=1}^{N}$ denote the set of $N$ category names. Each category label $c_i$ is encoded using a learnable textual prompt $P_T$ combined with a template-based description, yielding the normalized label text embedding $\text{T}^{(i)}_{\text{lbl}}$:
\begin{equation}
\text{T}^{(i)}_{\text{lbl}} = E_T(P_T \oplus \text{``a photo of a \{}c_i\text{\}.''})
\end{equation}

The complete set of label text embeddings $\text{T}_{\text{lbl}}$ is thus defined as:
\begin{equation}
\text{T}_{\text{lbl}} = [\text{T}^{(1)}_{\text{lbl}}, \text{T}^{(2)}_{\text{lbl}}, \ldots, \text{T}^{(N)}_{\text{lbl}}] \in \mathbb{R}^{N \times d}
\end{equation}

To incorporate complementary semantic signals, we enhance each $\text{T}^{(i)}_{\text{lbl}}$ by integrating it with its corresponding semantically attribute-aligned text embedding $\hat{\text{T}}^{(i)}_{\text{atr}}$ obtained from the semantic alignment module. Specifically, the two embeddings are concatenated and passed through a multi-layer perceptron (\text{MLP}) to learn a joint projection:
\begin{equation}
\tilde{\text{Z}}^{(i)} = \text{MLP}([{\text{T}^{(i)}_{\text{lbl}}} ; {\hat{\text{T}}^{(i)}_{\text{atr}}}]) \in \mathbb{R}^{d}
\end{equation}
where \text{MLP} serves to nonlinearly fuse category-specific identity and cross-category semantics into a unified embedding space, $[\cdot \ ; \ \cdot]$ denotes vector concatenation along the embedding dimension.

However, directly using $\tilde{\text{Z}}^{(i)}$ may dilute the core identity encoded in the original label representation. 
To preserve essential category semantics while still benefiting from auxiliary attributes, we introduce a residual connection:
\begin{equation}
{\text{T}^{(i)}_{\text{enh}}} = (1 - \alpha) \cdot {\text{T}^{(i)}_{\text{lbl}}} + \alpha \cdot {\tilde{\text{Z}}^{(i)}}
\end{equation}
where $\alpha \in [0,1]$ is a hyperparameter that controls the balance between the label text embeddings ${\text{T}_{\text{lbl}}}^{(i)}$ and the enriched representation ${\tilde{\text{Z}}}^{(i)}$. This formulation ensures that the enhanced embedding maintains a strong grounding in the original label semantics while flexibly incorporating auxiliary knowledge from textual descriptions.

By applying this refinement process to all $N$ categories, we obtain the set of attribute-enhanced text embeddings $\text{T}_{\text{enh}}$:
\begin{equation}
\text{T}_{\text{enh}} = [{\text{T}^{(1)}_{\text{enh}}}, {\text{T}^{(2)}_{\text{enh}}}, \ldots, {\text{T}^{(N)}_{\text{enh}}}] \in \mathbb{R}^{N \times d}
\end{equation}

This enables effective knowledge transfer by creating enriched representations that retain category-specific semantics while incorporating shared visual attributes. The residual formulation ensures stable training and prevents the dilution of essential categorical information, making $\text{T}_{\text{enh}}$ particularly effective for transfer learning where attribute-level knowledge is crucial for generalizing to unseen categories.

\subsection{Unified Training-Inference Adaptation}

To reconcile the dual goals of category-specific discrimination and cross-category generalization, we introduce a dual-embedding training and inference framework. It separates training objectives and aligns inference decisions by embedding type: label text embeddings $\text{T}_{\text{lbl}}$ are optimized for seen-category recognition, while attribute-enhanced text embeddings $\text{T}_{\text{enh}}$ encode transferable semantics for unseen-category generalization. This unified framework ensures task-aligned optimization during training and adaptive embedding selection during inference, supporting robust and flexible knowledge transfer across category boundaries.

\subsubsection{\textbf{Training Stage}}

The core challenge in generalization lies in learning representations that simultaneously excel at discriminating seen categories and encoding transferable knowledge for unseen ones. Traditional single-embedding approaches face an inherent trade-off: embeddings optimized for seen-category discrimination may lack the semantic richness needed for generalization. In contrast, embeddings enriched with attribute-level semantics may compromise discriminative precision. To address this dilemma, we propose a dual-embedding training strategy that decouples these competing objectives into complementary optimization paths.

We jointly optimize two complementary textual representations through separate classification losses, enabling the model to capture both precise category discrimination and rich cross-category semantic knowledge.

Specifically, the first loss operates on the label text embeddings $\text{T}_{\text{lbl}}$, encouraging sharp and distinct decision boundaries aligned with ground-truth categories:
\begin{equation}
\mathcal{L}_{\text{lbl}} = -\log \frac{\exp(\langle \text{v}, \text{T}_{\text{lbl}}^{(y)} \rangle / \tau)}{\sum_{j=1}^N \exp(\langle \text{v}, \text{T}_{\text{lbl}}^{(j)} \rangle / \tau)},
\end{equation}
where $y$ is the ground truth label, $\langle \cdot, \cdot \rangle$ denotes cosine similarity, and $\tau$ is the temperature parameter.

Simultaneously, to incorporate richer semantic relationships that transcend individual categories, we optimize attribute-enhanced text embeddings $\text{T}_{\text{enh}}$, which encode shared and transferable semantic attributes across categories:
\begin{equation}
\mathcal{L}_{\text{enh}} = -\log \frac{\exp(\langle \text{v}, \text{T}_{\text{enh}}^{(y)} \rangle / \tau)}{\sum_{j=1}^N \exp(\langle \text{v}, \text{T}_{\text{enh}}^{(j)} \rangle / \tau)}.
\end{equation}

This dual-objective design ensures that $\text{T}_{\text{lbl}}$ specializes in maximizing inter-category separation for precise classification of familiar categories, while $\text{T}_{\text{enh}}$ learns to capture shared semantic attributes that facilitate knowledge transfer to unseen categories. The independent optimization paths prevent mutual interference between discriminative and transferable objectives.

The overall training objective combines both losses with adaptive weighting:
\begin{equation}
\mathcal{L} = (1 - \beta) \cdot \mathcal{L}_{\text{lbl}} + \beta \cdot \mathcal{L}_{\text{enh}}
\end{equation}
where $\beta \in [0,1]$ controls the balance between discriminative precision and transferable semantics.

During training, we optimize only the learnable parameters while keeping the backbone CLIP encoders frozen. Specifically, the learnable parameters include the visual prompt $P_V$, the textual prompt $P_T$, and the linear projection layer \text{MLP} in the feature enhancement module. The frozen backbone encoders ($E_V$ and $E_T$) preserve the pre-trained cross-modal alignment knowledge.

\subsubsection{\textbf{Inference Stage}}

The core challenge in inference lies in making category predictions that preserve discriminative accuracy for seen categories while enabling reliable generalization to unseen ones. Traditional approaches often rely on a single embedding space for all categories, which either favors seen-category discrimination or overemphasizes semantic generalization, leading to suboptimal performance. To overcome this limitation, we introduce a dual-embedding inference strategy that adaptively selects between label-based and attribute-enhanced embeddings based on category exposure, thus aligning the decision mechanism with the nature of each category.

During inference, we employ an adaptive embedding selection strategy that dynamically assigns category-specific text embeddings based on whether a category has been seen during training. Specifically, the text embedding $\text{T}^{(i)}_{\text{inf}}$ used for category $c_i$ during inference is defined as:
\begin{equation}
{\text{T}^{(i)}_{\text{inf}}} =
\begin{cases}
{\text{T}^{(i)}_{\text{lbl}}}, & \text{if } c_i \in \mathcal{C}_s, \\
{\text{T}^{(i)}_{\text{enh}}}, & \text{if } c_i \in \mathcal{C}_u.
\end{cases}
\end{equation}
where ${\text{T}_{\text{inf}}}^{(i)}$ denotes the category-specific text embedding used for inference.

Therefore, the predicted category $\hat{y}$ is determined by:
\begin{equation}
\hat{y} = \arg\max_{c_i \in \mathcal{C}_{s} \cup \mathcal{C}_{u}} \langle{\text{v}}, {\text{T}^{(i)}_{\text{inf}}}\rangle
\end{equation}

This adaptive inference strategy leverages the complementary strengths of the two embedding types to support accurate and generalizable predictions. For seen categories, the label text embeddings $\text{T}_{\text{lbl}}$ optimized through direct visual supervision maintain sharp decision boundaries and strong discriminative power. In contrast, for unseen categories, the attribute-enhanced embeddings $\text{T}_{\text{enh}}$ enable effective knowledge transfer by capturing shared semantic structures across categories, thus compensating for the absence of direct supervision.

\section{Experiments}

\begin{table*}[!hbt]
\small
\centering
\caption{Comparison between our method and other existing methods under the Base-to-Novel Generalization setting.}
\label{base_to_novel}
\resizebox{\textwidth}{!}{
    \begin{tabular}{c|ccc|ccc|ccc|ccc}
    \toprule

    \multirow{2}{*}{Method} & \multicolumn{3}{c|}{\textit{Average}} & \multicolumn{3}{c|}{ImageNet} & \multicolumn{3}{c|}{Caltech101} & \multicolumn{3}{c}{OxfordPets} \\
     & Base & Novel & HM & Base & Novel & HM & Base & Novel & HM & Base & Novel & HM \\ \midrule
    $\text{CoOp}_{\text{ (IJCV2022)}}$ & 82.69 & 63.22 & 71.66 & 76.47 & 67.88 & 71.92 & 98.00 & 89.81 & 93.73 & 93.67 & 95.29 & 94.47 \\
    $\text{CoCoOp}_{\text{ (CVPR2022)}}$ & 80.47 & 71.69 & 75.83 & 75.98 & 70.43 & 73.10 & 97.96 & 93.81 & 95.84 & 95.20 &  97.69 & 96.43 \\
    $\text{MaPLe}_{\text{(CVPR2023)}}$ & 82.28 & 75.14 & 78.55 & 76.66 & 70.54 & 73.47 & 97.74 & 94.36 & 96.02 & 95.43 & 97.76 & 96.58 \\
    $\text{PromptSRC}_{\text{(ICCV2023)}}$ & 84.26 & 76.10 & 79.97 & 77.60 & 70.73 & 74.01 & 98.10 & 94.03 & 96.02 & 95.33 & 97.30 & 96.30 \\
    $\text{MetaPrompt}_{\text{(TIP2024)}}$ & 83.65 & 75.48 & 79.09 & 77.52 & 70.83 & 74.02 & 98.13 & 94.58 & 96.32 & 95.53 & 97.00 & 96.26 \\
    $\text{MMA}_{\text{(CVPR2024)}}$ & 83.20 & 76.80 & 79.87 & 77.31 & 71.00 & 74.02 & 98.40 & 94.00 & 96.15 & 95.40 & 98.07 & 96.72 \\
    $\text{KIM}_{\text{(TIP2024)}}$ & 84.70 & 76.10 & 80.20 & 77.90 & 69.00 & 73.20 & 98.40 & 92.60 & 95.40 & 94.70 & 97.40 & 96.00 \\
    $\text{PromptKD}_{\text{(CVPR2024)}}$ & 86.96 & 80.73 & 83.73 & \textbf{80.83} & 74.66 & 77.62 & 98.91 & 96.65 & \textbf{97.77} & 96.30 & 98.01 & 97.15 \\
    $\text{MMRL}_{\text{(CVPR2025)}}$ & 85.68 & 77.16 & 81.20 & 77.90 & 71.30 & 74.45 & 98.97 & 94.50 & 96.68 & 95.90 & 97.60 & 96.74 \\
    
    \midrule
    \multicolumn{1}{l|}{\footnotesize\textit{Using LLM:}} &\multicolumn{3}{c|}{} & \multicolumn{3}{c|}{} & \multicolumn{3}{c|}{} & \multicolumn{3}{c}{} \\
    $\text{HPT}_{\text{(AAAI2024)}}$ & 84.32 & 76.86 & 80.23 & 77.95 & 70.74 & 74.17 & 98.37 & 94.98 & 96.65 & 95.78 & 97.65 & 96.71 \\
    $\text{CoPrompt}_{\text{(ICLR2024)}}$ & 84.00 & 77.23 & 80.48 & 77.67 & 71.27 & 74.33 & 98.27 & 94.90 & 96.55 & 95.67 & 98.10 & 96.87 \\
    $\text{ArGue}_{\text{(CVPR2024)}}$ & 83.77 & 78.74 & 81.18 & 76.95 & 71.86 & 74.32 & 98.63 & 94.70 & 96.63 & 96.23 & 98.59 & 97.40 \\
    $\text{CoCoLe}_{\text{(ECCV2024)}}$ & 85.22 & 80.31 & 82.70 & 79.25 & 74.58 & 76.84 & 98.17 & 95.67 & 96.90 & 96.21 & 98.55 & 97.37 \\

    $\text{ProText}_{\text{(AAAI2025)}}$ & 72.95 & 76.98 & 74.91 & 75.00 & 71.38 & 73.14 & 98.06 & 95.63 & 96.83 & 94.95 & 98.00 & 96.45 \\
    
    \rowcolor{mygray}
    $\text{PromptKD \emph{w}/SemPT}_{\text{(Ours)}}$ & \textbf{87.08} & \textbf{81.21} & \textbf{84.04} & 80.71 & \textbf{74.82} & \textbf{77.65} & 98.77 & \textbf{96.77} & 97.76 & \textbf{96.38} & \textbf{98.37} & \textbf{97.36} \\

    \rowcolor{mygray}
    $\text{MMRL \emph{w}/SemPT}_{\text{(Ours)}}$ & 85.96 & 77.54 & 81.53 & 77.74 & 71.62 & 74.55 & \textbf{98.91} & 94.75 & 96.79 & 95.84 & 97.72 & 96.77 \\

    \midrule \midrule
    
    \multirow{2}{*}{Method} & \multicolumn{3}{c|}{StanfordCars} & \multicolumn{3}{c|}{Flowers102} & \multicolumn{3}{c|}{Food101} & \multicolumn{3}{c}{FGVCAircraft} \\
     & Base & Novel & HM & Base & Novel & HM & Base & Novel & HM & Base & Novel & HM \\ \midrule
    $\text{CoOp}_{\text{(IJCV2022)}}$ & 78.12 & 60.40 & 68.13 & 97.60 & 59.67 & 74.06 & 88.33 & 82.26 & 85.19 & 40.44 & 22.30 & 28.75 \\
    $\text{CoCoOp}_{\text{(CVPR2022)}}$ & 70.49 & 73.59 & 72.01 & 94.87 & 71.75 & 81.71 & 90.70 & 91.29 & 90.99 & 33.41 & 23.71 & 27.74 \\
    $\text{MaPLe}_{\text{(CVPR2023)}}$ & 72.94 & 74.00 & 73.47 & 95.92 & 72.46 & 82.56 & 90.71 & 92.05 & 91.38 & 37.44 & 35.61 & 36.50 \\
    $\text{PromptSRC}_{\text{(ICCV2023)}}$ & 78.27 & 74.97 & 76.58 & 98.07 & 76.50 & 85.95 & 90.67 & 91.53 & 91.10 & 42.73 & 37.87 & 40.15 \\
    $\text{MetaPrompt}_{\text{(TIP2024)}}$ & 76.34 & 75.01 & 75.48 & 97.66 & 74.49 & 84.52 & 90.74 & 91.85 & 91.29 & 40.14 & 36.51 & 38.24 \\
    $\text{MMA}_{\text{(CVPR2024)}}$ & 78.50 & 73.10 & 75.70 & 97.77 & 75.93 & 85.48 & 90.13 & 91.30 & 90.71 & 40.57 & 36.33 & 38.33 \\
    $\text{KIM}_{\text{(TIP2024)}}$ & 81.00 & 73.40 & 77.00 & 97.80 & 76.30 & 85.70 & 90.50 & 91.10 & 90.80 & 42.80 & 37.70 & 30.10 \\
    $\text{PromptKD}_{\text{(CVPR2024)}}$ & \textbf{82.80} & 83.37 & 83.13 & 99.42 & 82.62 & 90.24 & 92.43 & 93.68 & 93.05 & 49.12 & 41.81 & 45.17 \\
    $\text{MMRL}_{\text{(CVPR2025)}}$ & 81.30 & 75.07 & 78.06 & 98.97 & 77.27 & 86.78 & 90.57 & 91.50 & 91.03 & 46.30 & 37.03 & 41.15 \\
    
    \midrule
    \multicolumn{1}{l|}{\footnotesize\textit{Using LLM:}} &\multicolumn{3}{c|}{} & \multicolumn{3}{c|}{} & \multicolumn{3}{c|}{} & \multicolumn{3}{c}{} \\
    $\text{HPT}_{\text{(AAAI2024)}}$ & 76.95 & 74.23 & 75.57 & 98.17 & 78.37 & 87.16 & 90.46 & 91.57 & 91.01 & 42.68 & 38.13 & 40.28 \\
    $\text{CoPrompt}_{\text{(ICLR2024)}}$ & 76.97 & 74.40 & 75.66 & 97.27 & 76.60 & 85.71 & 90.73 & 92.07 & 91.40 & 40.20 & 39.33 & 39.76 \\   
    $\text{ArGue}_{\text{(CVPR2024)}}$ & 75.06 & 74.18 & 74.62 & 98.62 & 77.96 & 87.08 & 91.42 & 92.40 & 91.91 & 41.29 & 38.80 & 40.41 \\
    $\text{CoCoLe}_{\text{(ECCV2024)}}$ & 80.32 & 78.84 & 79.57 & 97.72 & 81.04 & 88.60 & 92.23 & 94.28 & 93.24 & 43.86 & 42.65 & 43.25 \\
    
    $\text{ProText}_{\text{(AAAI2025)}}$ & 64.54 & 76.08 & 68.84 & 74.36 & 78.44 & 76.35 & 90.20 & 91.98 & 91.08 & 30.91 & 34.13 & 32.44 \\
    
    \rowcolor{mygray}
    $\text{PromptKD \emph{w}/SemPT}_{\text{(Ours)}}$ & 82.73 & \textbf{84.30} & \textbf{83.51} & \textbf{99.52} & \textbf{82.75} & \textbf{90.36} & \textbf{92.50} & \textbf{93.74} & \textbf{93.12} & \textbf{49.40} & \textbf{42.35} & \textbf{45.60} \\

    \rowcolor{mygray}
    $\text{MMRL \emph{w}/SemPT}_{\text{(Ours)}}$ & 81.89 & 75.48 & 78.55 & 98.86 & 76.84 & 86.47 & 90.88 & 91.76 & 91.32 & 46.23 & 38.31 & 41.90 \\
      
    \midrule \midrule
      
    \multirow{2}{*}{Method} & \multicolumn{3}{c|}{SUN397} & \multicolumn{3}{c|}{DTD} & \multicolumn{3}{c|}{EuroSAT} & \multicolumn{3}{c}{UCF101} \\
     & Base & Novel & HM & Base & Novel & HM & Base & Novel & HM & Base & Novel & HM \\ \midrule
    $\text{CoOp}_{\text{(IJCV2022)}}$ & 80.60 & 65.89 & 72.51 & 79.44 & 41.18 & 54.24 & 92.19 & 54.74 & 68.69 & 84.69 & 56.05 & 67.46 \\
    $\text{CoCoOp}_{\text{(CVPR2022)}}$ & 79.74 & 76.86 & 78.27 & 77.01 & 56.00 & 64.85 & 87.49 & 60.04 & 71.21 & 82.33 & 73.45 & 77.64 \\
    $\text{MaPLe}_{\text{(CVPR2023)}}$ & 80.82 & 78.70 & 79.75 & 80.36 & 59.18 & 68.16 & 94.07 & 73.23 & 82.35 & 83.00 & 78.66 & 80.77 \\
    $\text{PromptSRC}_{\text{(ICCV2023)}}$ & 82.67 & 78.47 & 80.52 & 83.37 & 62.97 & 71.75 & 92.90 & 73.90 & 82.32 & 87.10 & 78.80 & 82.74 \\
    $\text{MetaPrompt}_{\text{(TIP2024)}}$ & 82.26 & 79.04 & 80.62 & 83.10 & 58.05 & 68.35 & 93.53 & 75.21 & 83.38 & 85.33 & 77.72 & 81.35 \\
    $\text{MMA}_{\text{(CVPR2024)}}$ & 82.27 & 78.57 & 80.38 & 83.20 & 65.63 & 73.38 & 85.46 & 82.34 & 83.87 & 86.23 & 80.03 & 82.20 \\
    $\text{KIM}_{\text{(TIP2024)}}$ & 83.10 & 78.50 & 80.7 & 84.10 & 66.20 & 74.10 & 94.80 & 79.00 & 86.20 & 86.30 & 75.90 & 80.80 \\
    $\text{PromptKD}_{\text{(CVPR2024)}}$ & 83.69 & 81.54 & 82.60 & 85.84 & 71.37 & 77.94 & \textbf{97.54} & 82.08 & 89.14 & \textbf{89.71} & 82.27 & \textbf{86.10} \\
    $\text{MMRL}_{\text{(CVPR25)}}$ & 83.20 & 79.30 & 81.20 & 85.67 & 65.00 & 73.82 & 95.60 & 80.17 & 87.21 & 88.10 & 80.07 & 83.89 \\
    
    \midrule
    \multicolumn{1}{l|}{\footnotesize\textit{Using LLM:}} &\multicolumn{3}{c|}{} & \multicolumn{3}{c|}{} & \multicolumn{3}{c|}{} & \multicolumn{3}{c}{} \\
    $\text{HPT}_{\text{(AAAI2024)}}$ & 82.57 & 79.26 & 80.88 & 83.84 & 63.33 & 72.16 & 94.24 & 77.12 & 84.82 & 86.52 & 80.06 & 83.16 \\
    $\text{CoPrompt}_{\text{(ICLR2024)}}$ & 82.63 & 80.03 & 81.31 & 83.13 & 64.73 & 72.79 & 94.60 & 78.57 & 85.84 & 86.90 & 79.57 & 83.07 \\    
    $\text{ArGue}_{\text{(CVPR2024)}}$ & 81.89 & 80.48 & 81.18 & 80.33 & 67.03 & 73.08 & 95.10 & 90.68 & 92.84 & 86.00 & 79.43 & 82.58 \\
    $\text{CoCoLe}_{\text{(ECCV2024)}}$ & 83.97 & 82.24 & 83.10 & 82.46 & 68.38 & 74.76 & 95.03 & 84.17 & 89.27 & 88.30 & 83.05 & 85.60 \\

    $\text{ProText}_{\text{(AAAI2025)}}$ & 76.14 & 79.14 & 77.61 & 63.08 & 61.59 & 62.33 & 59.71 & 80.97 & 68.73 & 75.54 & 79.50 & 77.47 \\
    
    \rowcolor{mygray}
    $\text{PromptKD \emph{w}/SemPT}_{\text{(Ours)}}$ & \textbf{83.81} & \textbf{81.69} & \textbf{82.74} & \textbf{87.03} & \textbf{71.62} & \textbf{78.58} & 97.40 & \textbf{84.46} & \textbf{90.47} & 89.62 & \textbf{82.47} & 85.90 \\

    \rowcolor{mygray}
    $\text{MMRL \emph{w}/SemPT}_{\text{(Ours)}}$ & 83.27 & 79.69 & 81.44 & 86.69 & 65.22 & 74.44 & 97.29 & 80.77 & 88.26 & 87.94 & 80.82 & 84.23 \\
    
    \bottomrule
    \end{tabular}
    }
\vspace{-0.3cm}
\end{table*}

\subsection{Datasets}
We evaluate our approach on 15 diverse datasets, which are categorized into two functional groups according to their roles in the evaluation: the Standard Benchmarks and the ImageNet Series.
The Standard Benchmarks group consists of 11 widely-used datasets that span a broad range of recognition tasks and visual domains. These include general object classification (e.g., ImageNet~\cite{ImageNet}, Caltech101~\cite{Caltech101}), fine-grained recognition (e.g., StanfordCars~\cite{StanfordCars}, OxfordPets~\cite{OxfordPets}, FGVCAircraft~\cite{FGVCAircraft}, Flowers102~\cite{Flowers102}, Food101~\cite{Food101}), scene understanding (SUN397~\cite{SUN397}), texture classification (DTD~\cite{DTD}), remote sensing (EuroSAT~\cite{EuroSAT}), and human action recognition (UCF101~\cite{UCF101}). These datasets serve as the primary testbed for assessing generalization across various visual domains and task types under standard conditions.
The ImageNet Series consists of ImageNet and four challenging variants: ImageNet-V2~\cite{ImageNet-V2}, ImageNet-Sketch~\cite{ImageNet-S}, ImageNet-A~\cite{ImageNet-A}, and ImageNet-R~\cite{ImageNet-R}. These datasets share the same label space as ImageNet but introduce significant distribution shifts. This group is specifically curated to evaluate model robustness and domain transferability under controlled perturbations and real-world distributional changes.

\begin{table*}[h]
\centering
\caption{Comparison between our method and other existing methods under the Cross-Dataset Transfer setting.}
\label{cross_dataset}
{
\resizebox{\textwidth}{!}{
    \begin{tabular}{c|c|ccccccccccc}
    \toprule
    \multirow{2}{*}{\textbf{Method}} & \textbf{Source} & \multicolumn{11}{c}{\textbf{Target}} \\
    \cmidrule(lr){2-13} & ImNet & Caltech & Pets & Cars & Flowers & Food & Aircraft & SUN & DTD & EuroSAT & UCF & \textit{Avg.} \\
    \midrule 
    $\text{CoOp}_{\text{ (IJCV2022)}}$ & 71.51 & 93.70 & 89.14 & 64.51 & 68.71 & 85.30 & 18.47 & 64.15 & 41.92 & 46.39 & 66.55 & 63.88 \\
    $\text{CoCoOp}_{\text{ (CVPR2022)}}$ & 71.02 & 94.43 & 90.14 & 65.32 & 71.88 & 86.06 & 22.94 & 67.36 & 45.73 & 45.37 & 68.21 & 65.74 \\
    $\text{MaPLe}_{\text{ (CVPR2023)}}$ & 70.72 & 93.53 & 90.49 & 65.57 & 72.23 & 86.20 & 24.74 & 67.01 & 46.49 & 48.06 & 68.69 & 66.30 \\
    $\text{PromptSRC}_{\text{ (ICCV2023)}}$ & 71.27 & 93.60 & 90.25 & 65.70 & 70.25 & 86.15 & 23.90 & 67.10 & 46.87 & 45.50 & 68.75 & 65.81 \\
    $\text{MMA}_{\text{ (CVPR2024)}}$ & 71.00 & 93.80 & 90.30 & 66.13 & 72.07 & 86.12 & 25.33 & 68.17 & 46.57 & 49.24 & 68.32 & 66.61 \\ 
    $\text{KIM}_{\text{ (TIP2024)}}$ & 73.52 & 94.05 & 90.65 & 64.20 & 71.10 & 85.60 & 25.20 & 67.10 & 45.25 & 51.45 & 68.60 & 66.32 \\ 
    $\text{PromptKD}_{\text{ (CVPR2024)}}$ & 72.42 & 93.61 & 91.59 & 73.93 & 75.33 & 88.84 & 26.24 & 68.57 & 55.08 & 63.74 & \textbf{76.39} & 71.33 \\ 
    $\text{MMRL}_{\text{ (CVPR2025)}}$ & 72.03 & 94.67 & 91.43 & 66.10 & 72.77 & 86.40 & 26.30 & 67.57 & 45.90 & 53.10 & 68.27 & 67.25 \\ 
    \midrule
    \multicolumn{1}{l|}{\footnotesize\textit{Using LLM:}} &\multicolumn{1}{c|}{} & \multicolumn{11}{c}{} \\

    $\text{HPT}_{\text{(AAAI2024)}}$ & 71.72 & 94.20 & \textbf{92.63} & 66.33 & 74.84 & 86.21 & 25.68 & 68.75 & 50.87 & 47.36 & 70.50 & 67.74 \\
    $\text{CoPrompt}_{\text{(ICLR2024)}}$ & 70.80 & 94.50 & 90.73 & 65.97 & 72.30 & 86.43 & 24.00 & 67.57 & 47.07 & 51.90 & 69.73 & 67.00 \\    
    $\text{CoCoLe}_{\text{(ECCV2024)}}$ & \textbf{73.88} & \textbf{95.88} & 91.93 & 67.79 & 74.17 & 87.97 & \textbf{28.83} & 68.75 & 49.26 & 51.75 & 72.78 & 68.91 \\
    
    $\text{ProText}_{\text{(AAAI2025)}}$ & 69.80 & 94.81 & 91.01 & 66.60 & 72.35 & 86.66 & 24.72 & 67.34 & 47.93 & 51.86 & 69.60 & 67.23 \\
    
    \rowcolor{mygray}
    $\text{PromptKD \emph{w}/SemPT}_{\text{ (Ours)}}$ & 72.56 & 93.92 & 92.17 & \textbf{75.24} & \textbf{75.52} & \textbf{89.31} & 28.26 & \textbf{68.76} & \textbf{57.03} & \textbf{65.23} & 75.64 & \textbf{72.11} \\ 

    \rowcolor{mygray}
    $\text{MMRL \emph{w}/SemPT}_{\text{ (Ours)}}$ & 72.21 & 93.97 & 91.94 & 66.83 & 73.32 & 86.64 & 27.96 & 67.67 & 46.91 & 54.45 & 69.04 & 67.87 \\ 
    \bottomrule
    \end{tabular}
    }
}
\vspace{-0.3cm}
\end{table*}

\subsection{Evaluation Settings}
We evaluate our approach across four distinct settings, each designed to examine a different aspect of generalization. These settings are selected to provide a comprehensive assessment of the model’s performance under varying levels of supervision and distributional shift.

\subsubsection{\textbf{Base-to-Novel Generalization}}
In the Base-to-Novel Generalization setting, the categories within each dataset are evenly split into base and novel subsets. The model is trained solely on the base categories and evaluated on both base and novel categories without any fine-tuning. This evaluation measures the model’s ability to retain discriminative power on seen categories while generalizing to previously unseen ones. It reflects the model’s capacity to transfer learned representations for recognizing new concepts, thereby balancing specialization and generalization.

\subsubsection{\textbf{Cross-Dataset Transfer}}
In the Cross-Dataset Transfer setting, the model is trained on one dataset and directly evaluated on a set of entirely different target datasets, without any adaptation or fine-tuning. This evaluation rigorously measures the model’s capacity to learn domain-agnostic and transferable image representations that can generalize effectively across datasets with diverse data distributions, label spaces, and visual domains. It provides critical insight into the model’s robustness and adaptability when deployed in previously unseen environments.

\subsubsection{\textbf{Cross-Domain Transfer}}
In the Cross-Domain Transfer setting, the model is trained on the original dataset and evaluated on several variants without any further adaptation. These variants share the same label space as ImageNet but introduce significant visual shifts in style, texture, or composition. This evaluation focuses on the model's resilience to natural distribution shifts while maintaining consistency in category recognition, which is crucial for real-world deployment under changing visual conditions. It demonstrates the model's ability to maintain robust performance when encountering visual variations that commonly occur in practical applications.

\subsubsection{\textbf{Few-Shot Learning}}

In the Few-Shot Learning setting, the model is trained with only a limited number of labeled examples per category and evaluated on the whole test set. To simulate varying levels of data scarcity, we consider five shot configurations: 1, 2, 4, 8, and 16 examples per category. This evaluation directly examines the model’s sample efficiency and its ability to generalize from minimal supervision. It reveals how effectively the model can leverage prior knowledge and semantic understanding to achieve strong performance even when training data is severely limited.

We conduct Base-to-Novel Generalization, Cross-Dataset Transfer, and Few-Shot Learning settings on the Standard Benchmarks, while the Cross-Domain Transfer setting is performed on the ImageNet Series.

\subsection{Implementation Details}

All experiments are implemented using the PyTorch~\cite{PyTorch} framework and conducted on NVIDIA RTX 3090 GPUs. For our approach, we set the key hyperparameters as follows: the number of top-ranked attribute text embeddings for similarity-based selection is fixed at $K=2$ across all experiments. The balancing coefficient $\alpha$ in Equation 10, which controls the contribution of our selective textual feature integration, is set to 0.2. Additionally, the weighting factor $\beta$ in Equation 14, which regulates the influence of the proposed feature alignment objective, is set to 0.4. For baseline integration, we use PromptKD~\cite{PromptKD} and MMRL~\cite{MMRL} with our SemPT for Base-to-Novel Generalization, Cross-Dataset Transfer, and Cross-Domain Transfer experiments, while Few-Shot Learning experiments use MMRL with SemPT. For fair comparison, all baseline methods were faithfully reproduced by strictly adhering to their original parameter configurations, including the number of training epochs, batch size, learning rate, optimizer choice, and so on. This ensures that performance differences can be attributed to our proposed methods rather than implementation variations.

\subsection{Experimental Results}

In this section, we present and analyze the experimental results of our proposed approach across the four different evaluation tasks. We compare our method with state-of-the-art baselines and provide detailed performance analyses.

\begin{table}[t]
\centering
\caption{Comparison between our method and other existing methods under the Cross-Domain Transfer setting.}
\label{cross_domain}
\resizebox{0.5\textwidth}{!}{
    \footnotesize
    \begin{tabular}{c|c|ccccc}
    \toprule
    \multirow{2}{*}{\textbf{Method}} & \textbf{Source} & \multicolumn{5}{c}{\textbf{Target}} \\
    \cmidrule(lr){2-7}                             & ImageNet & -V2 & -S & -A & -R & \textit{Avg.} \\
    \midrule
    $\text{CoOp}_{\text{ (IJCV2022)}}$   & 71.51    & 64.20 & 47.99 & 49.71 & 75.21 & 59.28 \\
    $\text{CoCoOp}_{\text{ (CVPR2022)}}$ & 71.02    & 64.07 & 48.75 & 50.63 & 76.18 & 59.91 \\
    $\text{MaPLe}_{\text{ (CVPR2023)}}$  & 70.72    & 64.07 & 49.15 & 50.90 & 76.98 & 60.27 \\
    $\text{PromptSRC}_{\text{ (ICCV2023)}}$ & 71.27 & 64.35 & 49.55 & 50.90 & 77.80 & 60.65 \\
    $\text{MMA}_{\text{ (CVPR2024)}}$    & 71.00    & 64.33 & 49.13 & 51.12 & 77.32 & 60.48 \\ 
    $\text{KIM}_{\text{ (TIP2024)}}$    & 73.52    & 64.76 & 48.84 & 49.80 & 76.72 & 60.03 \\
    $\text{PromptKD}_{\text{ (CVPR2024)}}$ & 72.42  & 69.77 & 58.72 & 70.36 & 87.01 & 71.47 \\ 
    $\text{MMRL}_{\text{ (CVPR2025)}}$       & 72.03 & 64.47 & 49.17 & 51.20 & 77.53 & 60.59 \\
    \midrule
    \multicolumn{1}{l|}{\footnotesize\textit{Using LLM:}} &\multicolumn{1}{c|}{} & \multicolumn{5}{c}{} \\

    $\text{HPT}_{\text{(AAAI2024)}}$ & 71.72 & 65.25 & 49.36 & 50.85 & 77.38 & 60.71 \\
    $\text{CoPrompt}_{\text{(ICLR2024)}}$ & 70.80 & 64.25 & 49.43 & 50.50 & 77.51 & 60.42 \\  
    $\text{ArGue}_{\text{(CVPR2024)}}$ & 71.84 & 65.02 & 49.25 & 51.47 & 76.96 & 60.68 \\
    $\text{CoCoLe}_{\text{(ECCV2024)}}$ & \textbf{73.88} & 65.86 & 50.89 & 51.75 & 78.89 & 61.85 \\
    
    $\text{ProText}_{\text{(AAAI2025)}}$ & 70.22 & 63.54 & 49.45 & 51.47 & 77.35 & 60.45 \\
    
    \rowcolor{mygray}
    $\text{PromptKD \emph{w}/SemPT}_{\text{ (Ours)}}$ & 72.56 & \textbf{70.46} & \textbf{59.48} & \textbf{70.83} & \textbf{87.42} & \textbf{72.05} \\ 
    \rowcolor{mygray}
    $\text{MMRL \emph{w}/SemPT}_{\text{ (Ours)}}$ & 72.21 & 65.14 & 49.97 & 51.87 & 77.94 & 61.23 \\ 
    \bottomrule
    \end{tabular}
}
\vspace{-0.3cm}
\end{table}

We evaluate our approach against two categories of baseline methods. The first category includes traditional prompt tuning approaches: CoOp~\cite{CoOp}, CoCoOp~\cite{CoCoOp}, MaPLe~\cite{MaPLe}, PromptSRC~\cite{PromptSRC}, MetaPrompt~\cite{MetaPrompt}, MMA~\cite{MMA}, KIM~\cite{KIM}, PromptKD~\cite{PromptKD}, and MMRL~\cite{MMRL}. The second category comprises methods that incorporate LLM: HPT~\cite{HPT}, CoPrompt~\cite{CoPrompt}, ArGue~\cite{ArGue}, CoCoLe~\cite{CoCoLe}, and ProText~\cite{ProText}. To ensure fair comparison with these LLM-based baseline methods, our approach follows the same experimental setting by leveraging LLM-generated descriptions for all categories.

\begin{figure*}[!hbt]
  \centering
  \includegraphics[width=\textwidth]{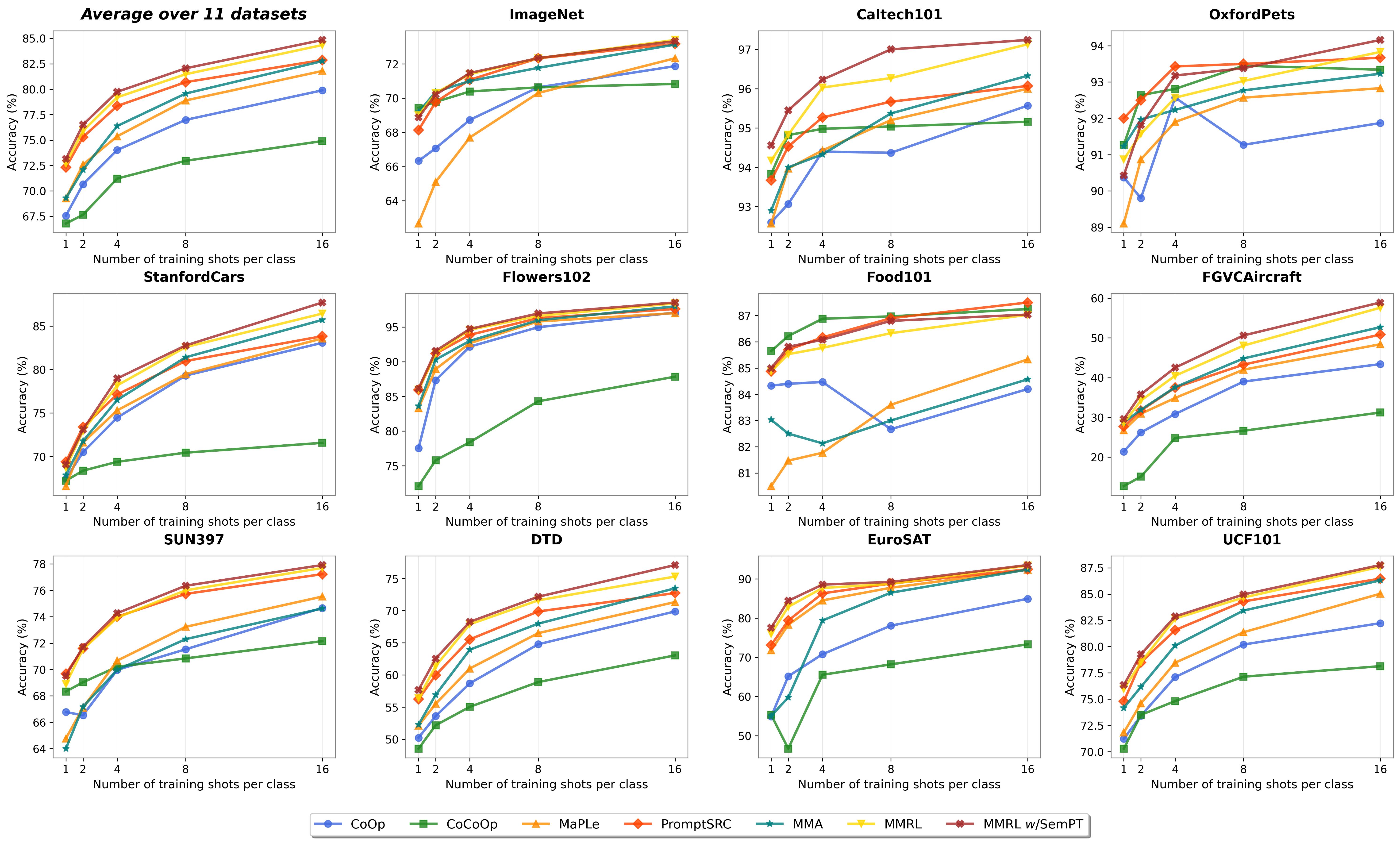}
  \caption{Comparison between our method and other existing methods under the Few-Shot Learning setting.}
  \label{fig:fewshot} 
\end{figure*}

\subsubsection{\textbf{Base-to-Novel Generalization Results}}

Table \ref{base_to_novel} presents comprehensive results comparing our SemPT approach with existing state-of-the-art methods across 11 datasets under the base-to-novel generalization setting. Our method demonstrates consistent improvements when integrated with strong baseline approaches. Specifically, PromptKD \emph{w}/SemPT achieves the best overall performance with an average harmonic mean (HM) of 84.04\%, representing a notable improvement of 0.31\% over the original PromptKD method. This enhancement is particularly significant given that PromptKD already represents one of the strongest baselines in this setting. Similarly, MMRL \emph{w}/SemPT shows improvements across most datasets, achieving competitive results while maintaining the robustness of the base method.

The experimental findings reveal several critical insights regarding the efficacy of our approach. First, our semantic prompt tuning methodology consistently augments performance across both base and novel categories, substantiating the hypothesis that LLM-generated semantic descriptions furnish semantically rich complementary information that enhances few-shot learning capabilities. Second, the performance gains are particularly salient on fine-grained classification datasets such as StanfordCars and OxfordPets, where nuanced semantic descriptions effectively capture subtle inter-category visual distinctions that are imperative for precise categorization. Third, when juxtaposed with other LLM-enhanced methodologies, including HPT, CoPrompt, and CoCoLe, our approach demonstrates superior empirical performance across the evaluated benchmarks. The consistent improvements observed across heterogeneous datasets substantiate the generalizability and robustness of our semantic prompt tuning paradigm in effectively mitigating the domain gap between base and novel recognition tasks.

\subsubsection{\textbf{Cross-Dataset Transfer Results}}

Table \ref{cross_dataset} presents a comprehensive evaluation of our SemPT methodology under the cross-dataset transfer paradigm, wherein models are exclusively trained on ImageNet and subsequently assessed across 11 heterogeneous target datasets without any domain-specific adaptation. Our approach demonstrates compelling performance enhancements across the evaluated benchmarks, with PromptKD \emph{w}/SemPT establishing a new state-of-the-art average accuracy of 72.11\%, constituting a substantial improvement of 0.78\% over the original PromptKD framework. This advancement is particularly noteworthy given the inherent complexity of cross-dataset generalization, where pronounced domain gaps and distributional divergences present formidable challenges to model transferability.

The empirical findings illuminate several critical aspects of cross-domain generalizability facilitated by our semantic prompt tuning paradigm. Notably, our methodology exhibits particularly salient improvements on challenging fine-grained recognition tasks, exemplified by StanfordCars (75.24\% vs 73.93\%) and DTD (57.03\% vs 55.08\%), where the incorporation of semantically rich descriptions proves instrumental in disambiguating visually similar categories across disparate domains. Moreover, the consistent performance augmentation observed across diverse target domains, encompassing natural scene recognition (SUN397), aerial imagery classification (EuroSAT), and specialized aviation datasets (FGVCAircraft), substantiates the robustness and versatility of our approach. The semantic knowledge distilled from attribute-level descriptions appears to furnish domain-agnostic representations that effectively bridge the semantic gap between source and target distributions, thereby enhancing cross-domain discriminative capacity while mitigating the deleterious effects of domain shift.

\begin{table*}[ht]
\centering
\caption{Ablation Study for Various Components}
\label{tab:ablation}

\begin{minipage}{0.49\textwidth}
    \centering
    (a) Text Embedding Choice
    \begin{tabular}{l|ccc}
    \hline
    Method & Base & Novel & HM \\
    \hline
    $\text{T}^{\text{lbl}}$ for all categories & 87.08 & 80.56 & 83.69 \\
    $\text{T}^{\text{enh}}$ for all categories & 86.78 & 81.21 & 83.91 \\
    \rowcolor{mygray}
    \makecell[l]{\cellcolor{mygray}$\text{T}^{\text{lbl}}$ for seen categories \\ \cellcolor{mygray}$\text{T}^{\text{enh}}$ for unseen categories} & \textbf{87.08} & \textbf{81.21} & \textbf{84.04} \\
    \hline
    \end{tabular}
\end{minipage}%
\hfill
\begin{minipage}{0.49\textwidth}
    \centering
    (b) Knowledge Construction Strategy
    \begin{tabular}{l|ccc}
    \hline
    Strategy & Base & Novel & HM \\
    \hline
    Using category names only & 86.92 & 80.95 & 83.83 \\
    One-step prompting strategy & 86.86 & 80.85 & 83.75 \\
    \rowcolor{mygray}
    Two-step prompting strategy & \textbf{87.08} & \textbf{81.21} & \textbf{84.04} \\
    \hline
    \end{tabular}
\end{minipage}

\vspace{0.5cm}

\begin{minipage}{0.32\textwidth}
    \centering
    (c) Hyperparameter $\alpha$
    \begin{tabular}{c|ccc}
    \hline
    $\alpha$ & Base & Novel & HM \\
    \hline
    \rowcolor{mygray} 0.2 & \textbf{87.08} & \textbf{81.21} & \textbf{84.04} \\
    0.4 & 86.53 & 80.92 & 83.63 \\
    0.6 & 86.75 & 80.75 & 83.64 \\
    0.8 & 86.72 & 80.96 & 83.74 \\
    \hline
    \end{tabular}
\end{minipage}%
\hfill
\begin{minipage}{0.32\textwidth}
    \centering
    (d) Hyperparameter $\beta$
    \begin{tabular}{c|ccc}
    \hline
    $\beta$ & Base & Novel & HM \\
    \hline
    0.2 & 86.92 & 80.95 & 83.83 \\
    0.4 & 86.86 & 80.85 & 83.75 \\
    \rowcolor{mygray} 0.6 & \textbf{87.08} & \textbf{81.21} & \textbf{84.04} \\
    0.8 & 86.81 & 80.89 & 83.75 \\
    \hline
    \end{tabular}
\end{minipage}%
\hfill
\begin{minipage}{0.32\textwidth}
    \centering
    (e) Hyperparameter $K$
    \begin{tabular}{c|ccc}
    \hline
    $K$ & Base & Novel & HM \\
    \hline
    1 & 87.01 & 80.74 & 83.76 \\
    \rowcolor{mygray} 2 & \textbf{87.08} & \textbf{81.21} & \textbf{84.04} \\
    3 & 86.59 & 80.93 & 83.66 \\
    \hline
    \end{tabular}
\end{minipage}

\end{table*}

\subsubsection{\textbf{Cross-Domain Transfer Results}}

Table \ref{cross_domain} evaluates the robustness of our SemPT approach under the cross-domain transfer setting, where models trained on ImageNet are directly transferred to four challenging ImageNet variants: ImageNet-V2, ImageNet-Sketch (-S), ImageNet-Adversarial (-A), and ImageNet-Rendition (-R). These variants present distinct domain shifts, including distribution changes, stylistic variations, adversarial perturbations, and artistic renditions, thereby constituting a rigorous testbed for assessing cross-domain generalizability. Our methodology demonstrates consistent improvements across all target domains, with PromptKD \emph{w}/SemPT achieving the highest average performance of 72.05\%, representing a meaningful enhancement of 0.58\% over the original PromptKD baseline.

The experimental results underscore the efficacy of semantic prompt tuning in mitigating domain shift challenges across diverse visual distributions. Particularly noteworthy is the performance on ImageNet-Sketch (59.48\% vs 58.72\%) and ImageNet-Adversarial (70.83\% vs 70.36\%), where the incorporation of rich semantic descriptions proves instrumental in maintaining discriminative capabilities despite substantial stylistic and adversarial perturbations. The consistent improvements observed across all ImageNet variants, each presenting unique distributional challenges, demonstrate that LLM-generated semantic knowledge provides robust domain-invariant representations that effectively bridge the gap between natural and synthetic visual domains. Furthermore, the relatively more minor but consistent gains achieved by MMRL \emph{w}/SemPT (61.23\% vs 60.59\%) corroborate the general applicability of our semantic enhancement strategy across different baseline architectures. These findings substantiate the hypothesis that semantically enriched prompts facilitate more robust feature representations that maintain their discriminative power across diverse domain shifts, thereby enhancing the model's capacity for cross-domain generalization.

\subsubsection{\textbf{Few-Shot Learning Results}}

Fig.~\ref{fig:fewshot} presents the few-shot learning performance across different shot numbers (1, 2, 4, 8, 16) on 11 individual datasets as well as their averaged results. Our SemPT method consistently outperforms baseline approaches across different shot settings. Notably, our approach achieves the highest average accuracy of 84.84\% in the 16-shot setting, surpassing the strongest baseline MMRL by 0.5\%. The performance gains are particularly evident in challenging datasets such as FGVCAircraft and DTD, where the rich semantic descriptions generated by LLM provide crucial discriminative information for fine-grained recognition tasks. The consistent improvements across all shot numbers demonstrate that our semantic enhancement strategy effectively leverages limited training data, making it especially valuable for practical few-shot scenarios where labeled data is scarce.

\subsection{Ablation Studies}

Experiments are conducted under the base-to-novel generalization setting to evaluate each component's contribution. Our SemPT method serves as a plug-in component, and all experiments are conducted by integrating SemPT with PromptKD as the baseline framework.

\textbf{The Effect of Text Embedding Choice.} Table~\ref{tab:ablation}(a) compares three text embedding strategies for different category types. Our mixed strategy (label text embeddings $\text{T}^{\text{lbl}}$ for seen categories, attribute-enhanced text embeddings $\text{T}^{\text{enh}}$ for unseen categories) achieves the best performance: 87.08\% Base, 81.21\% Novel, and 84.04\% HM. Using attribute-enhanced text embeddings for all categories improves novel performance but slightly hurts base accuracy (86.78\% vs 87.08\%). The results demonstrate that label text embeddings are more effective for seen categories, as they have been specifically optimized for discriminative alignment through direct visual supervision during training. In contrast, unseen categories benefit significantly from enhanced embeddings, which encode transferable semantic attributes that facilitate cross-category knowledge transfer in the absence of visual training data. These findings support our adaptive embedding selection strategy, showing that different embedding types are optimal for different inference scenarios based on their specialized training objectives.

\textbf{The Effect of Knowledge Construction Strategy.} Table~\ref{tab:ablation}(b) compares three knowledge construction approaches. Two-step prompting strategy achieves the best results (84.04\% HM), outperforming one-step prompting strategy (83.75\% HM) and category names only (83.83\% HM). The superior performance of the two-step approach demonstrates the effectiveness of structured knowledge acquisition. The first step enables the LLM to establish a comprehensive understanding of each category's general information, while the second step facilitates more focused and discriminative description generation. This progressive refinement process ensures that the generated descriptions are both semantically rich and category-specific, mitigating potential ambiguity that may arise from single-step generation. These results highlight the importance of structured knowledge acquisition, where the two-step refinement process proves more effective than both single-step generation and using category names alone.

\textbf{The Effect of $\boldsymbol{\alpha}$.}
Table~\ref{tab:ablation}(c) shows the impact of fusion weight $\alpha$. Optimal performance occurs at $\alpha = 0.2$ (84.04\% HM), , with performance declining as $\alpha$ increases to 0.4 (83.63\% HM), 0.6 (83.64\% HM), and 0.8 (83.74\% HM). The low optimal $\alpha$ value reveals a critical insight about feature fusion in base-to-novel transfer learning. The original text embeddings, learned through extensive pre-training and fine-tuning processes, contain stable and reliable semantic representations that should maintain dominance in the final feature space. The LLM-generated text embeddings, while providing valuable supplementary information, should serve as a subtle augmentation rather than a replacement. When $\alpha$ becomes too large, the enhanced text embeddings may introduce noise or inconsistent semantic signals that conflict with the well-established original representations, leading to performance degradation.

\textbf{The Effect of $\boldsymbol{\beta}$.} Table~\ref{tab:ablation}(d) examines loss balancing weight $\beta$. The optimal $\beta$ = 0.6 achieves 84.04\% HM, outperforming lower values (0.2: 83.83\% HM, 0.4: 83.75\% HM) and higher values (0.8: 83.75\% HM). The fact that $\beta = 0.6$ outperforms lower values suggests that the model requires stronger supervision signals to learn how to utilize LLM-enhanced representations effectively. During training, the enhanced loss serves as a crucial guidance mechanism that teaches the model to leverage the enriched semantic space provided by attribute-level descriptions properly. However, when $\beta$ becomes too high (0.8), the model may over-emphasize the enhanced embeddings at the expense of maintaining good performance on seen categories, indicating the need for balanced supervision between standard and enhanced learning objectives.

\textbf{The Effect of $\boldsymbol{K}$.}  
Table~\ref{tab:ablation}(e) reports the results of varying the top-$K$ selection strategy. Setting $K = 2$ achieves the best performance (84.04\% HM), outperforming both $K = 1$ (83.76\% HM) and $K = 3$ (83.66\% HM). This result highlights a trade-off between semantic diversity and noise sensitivity in the selection of attribute-level descriptions. When $K = 1$, the model relies solely on the most similar description, which may lead to insufficient semantic coverage and limit its ability to capture complementary attributes. In contrast, increasing $K$ to 3 incorporates less relevant descriptions, potentially introducing semantic noise that weakens alignment with visual features. The optimal choice of $K = 2$ provides a balanced representation by incorporating diverse yet relevant semantics, enhancing category-level understanding while preserving discriminative focus.

\section{Conclusion}
In this paper, we introduced Semantic Prompt Tuning (SemPT), a framework that systematically incorporates attribute-level knowledge from large language models to construct semantically aligned and transferable text embeddings for transfer learning.
First, SemPT employs a two-step prompting strategy to guide  LLM in extracting shared visual attributes and generating attribute-level descriptions, capturing transferable semantic cues beyond labels while ensuring coherent structure. Second, visually guided weighting is applied to the embeddings of these attribute-level descriptions, reducing noise from irrelevant attributes and integrating them with label text embeddings to form attribute-enhanced text embeddings. Third, image embeddings are jointly aligned with both label and attribute-enhanced text embeddings, balancing discrimination for seen categories and transferability to unseen ones. Finally, during inference, the model dynamically selects between standard label embeddings for seen categories and attribute-enhanced embeddings for unseen ones, enabling effective adaptation based on category exposure.
Extensive experiments on 15 benchmark datasets confirm the effectiveness and versatility of SemPT, achieving consistent state-of-the-art results under diverse evaluation protocols and demonstrating strong compatibility with a wide range of vision-language backbones.

\bibliography{ms}
\end{document}